\documentclass[sigconf]{acmart}

\usepackage{multirow}
\usepackage{booktabs}
\usepackage{algorithm}
\usepackage{algorithmic}
\usepackage{graphicx}
\usepackage{arydshln} 
\usepackage{amsmath}
\usepackage{amsthm}

\usepackage{subfloat}
\usepackage{subfig}

\usepackage{fancyhdr}
\AtBeginDocument{%
	}

\setcopyright{acmcopyright}
\copyrightyear{2018}
\acmYear{2018}
\acmDOI{XXXXXXX.XXXXXXX}

\acmPrice{15.00}
\acmISBN{978-1-4503-XXXX-X/18/06}

\renewcommand\footnotetextcopyrightpermission[1]{}

\begin{document}
	
	\title{C$^2$MOE: Consistency and Complementarity-guided Mixture of Experts for Incomplete Multimodal Emotion Learning}

	\author{Yuntao Shou}
	\affiliation{
		  \institution{Central South University of Forestry and Technology}
		  \city{Hunan}
		  \country{China}}
	\email{shouyuntao@stu.xjtu.edu.cn}
	
		\author{Tao Meng}
				\authornote{Corresponding Author}
	\affiliation{
		\institution{Central South University of Forestry and Technology}
		\city{Hunan}
		\country{China}}
	\email{mengtao@hnu.edu.cn}

		\author{Wei Ai}
	\affiliation{
		\institution{Central South University of Forestry and Technology
			}
		\country{Hunan}
	\country{China}}
	\email{aiwei@hnu.edu.cn}
	
			\author{Keqin Li}
	\affiliation{
		\institution{State University of New York, New Paltz, New York 12561, USA}
		\city{Xi'an}
		\country{China}}
	\email{lik@newpaltz.edu}

	\renewcommand{\shortauthors}{Shou et al.}
	
	\begin{abstract}
Recent advances in Multimodal Emotion Recognition in Conversations (MERC) highlight its reliance on complete multimodal inputs. However, real-world data often suffer from missing modalities due to transmission errors or user behavior, severely degrading model performance. Existing methods enhance robustness via cross-modal consistency learning but largely ignore modality complementarity, leading to biased reconstructions. To address this limitation, we propose C²MOE, a novel Consistency and Complementarity-guided Mixture of Experts framework for incomplete multimodal emotion learning. Our approach unifies representation learning and missing modality imputation within a principled information-theoretic framework. Specifically, multimodal knowledge is factorized into consistency and complementarity components via interaction-aware experts. Consistency is captured by maximizing cross-modal predictability, while complementarity is preserved by maximizing conditional entropy between modalities. Building upon this decomposition, C²MOE introduces a dual-branch prediction mechanism for robust imputation under missing modalities. The consistency branch aligns imputed features with the joint distribution by minimizing uncertainty, and the complementarity branch exploits modality-unique cues via entropy maximization. Finally, C²MOE employs a learnable reweighting module that dynamically assigns importance scores to each expert’s output, yielding a robust and adaptive fusion for imputation. Extensive experiments on multiple MERC benchmarks demonstrate that C²MOE consistently surpasses state-of-the-art methods across various missing-modality settings, validating its robustness and generalization.
	\end{abstract}

	
	
	\maketitle
	
\begin{figure}
	\centering
	\includegraphics[width=1\linewidth]{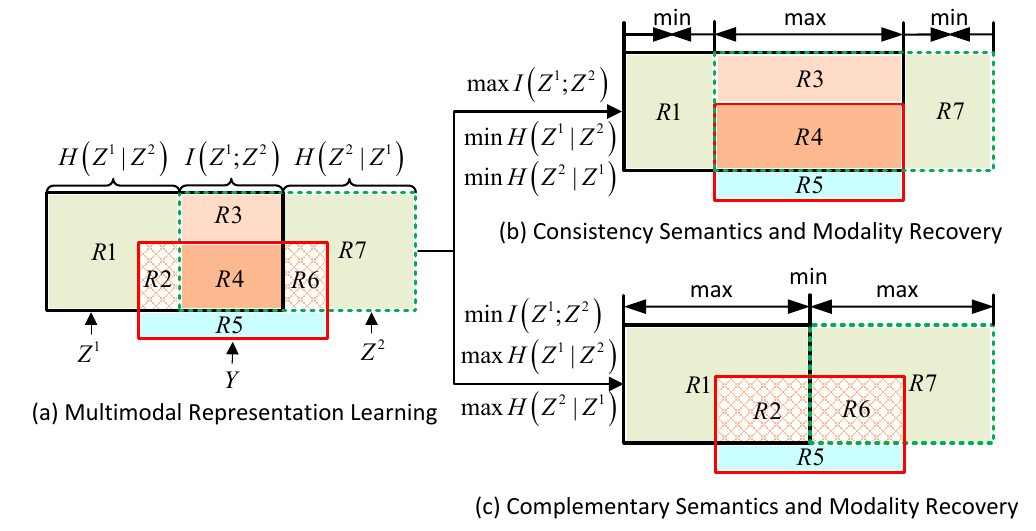}
	\vspace{-5mm}
	\caption{We present basic observations and theoretical insights from an information-theoretic perspective. (a) The black solid rectangle denotes the complete modality feature ${Z_1} = {R_1} \cup {R_2} \cup {R_3} \cup {R_4}$, the green dashed rectangle denotes the incomplete modality feature ${Z_2} = {R_3} \cup {R_4} \cup {R_6} \cup {R_7}$, and the red solid rectangle represents the task-related feature ${Y} = {R_2} \cup {R_4} \cup {R_5} \cup {R_6}$. Among them, ${R_1}$ and ${R_7}$ are modality-specific but task-irrelevant regions; ${R_2}$ and ${R_6}$ are modality-specific and task-relevant (complementary) regions; ${R_3}$ and ${R_4}$ are shared (consistent) regions, with ${R_3}$ being task-irrelevant and ${R_4}$ task-relevant; ${R_5}$ is a task-related but unknown region. (b) Maximizing consistent task-related semantics requires enlarging ${R_4}$ while suppressing ${R_1}$ and ${R_7}$. (c) Maximizing complementary task-related semantics, conversely, requires shrinking ${R_4}$ and expanding ${R_2}$ and ${R_6}$.}
	\label{fig:intro}
	\vspace{-5mm}
\end{figure}
\section{Introduction}


The core objective of multimodal emotion recognition in conversation (MERC) is to enhance emotion analysis accuracy and robustness by integrating heterogeneous modalities \cite{li2023decoupled, tsai2019multimodal, li2024toward, xu2024leveraging, shou2026comprehensive, shou2022conversational, shou2024adversarial, shou2024low}. Recent studies focus on learning discriminative cross-modal representations and developing effective fusion strategies \cite{ding2023lggnet, ramesh2021zero, hu2024novel, shou2025multimodal, meng2024deep, shou2025masked, meng2024multi, shou2025graph, shou2025cilf}. However, real-world scenarios often suffer from modality missing, which severely degrades model performance and robustness \cite{wang2023distribution}. Consequently, designing MERC models resilient to incomplete modalities has become a critical research challenge \cite{wang2024incomplete, zhang2024learning, liu2024contrastive, zhang2024towards, shou2026comprehensive, shou2025graph, shou2025multimodal, shou2026dual}.

For MERC under incomplete multimodality, a common solution is to use an encoder–decoder framework to recover missing modalities from observed ones. The encoder estimates the missing modality from available representations, and the decoder reconstructs it in the original feature space. For example, Zhao et al. \cite{zhao2021missing} combine autoencoders with cycle consistency to reduce distortion and noise, while Lian et al. \cite{lian2023gcnet} model multimodal features as graph nodes and propagate information with graph neural networks to complete missing modalities. However, these approaches offer limited interpretability of the recovery mechanism. To address this, information-theoretic methods \cite{lin2021completer, lin2022dual} enhance cross-modal consistency by maximizing mutual information between modalities, thereby improving performance under missing-modality conditions. Nevertheless, these methods ignore the impact of complementarity on multimodal learning.

However, jointly modeling inter-modal consistency and complementarity presents a fundamental challenge because the two objectives are inherently conflicting. Promoting consistency requires minimizing conditional entropy $H(\mathbf{Z}^i\mid\mathbf{Z}^j)$ to reduce cross-modal uncertainty, while enhancing complementarity relies on maximizing the same quantity to preserve modality-specific information. To address this, we propose the Consistency and Complementarity-guided Mixture of Experts framework (C$^2$MOE), which decouples the two objectives through specialized experts and adaptive routing. C$^2$MOE employs two experts within a mixture-of-experts architecture. The consistency expert maximizes mutual information $I(\mathbf{Z}^1, \mathbf{Z}^2)$ and minimizes conditional entropy to align recovered data with the true distribution. The complementarity expert performs the opposite optimization to retain unique modality features and strengthen reconstruction robustness. A dynamic gating network assigns each input to the appropriate expert based on modality availability and semantic context, and a coordination module fuses their outputs into a unified representation that balances consistency and complementarity. Overall, our contributions are as:
\begin{itemize}
	\item To the best of our knowledge, we make the first attempt to integrate consistency and complementary semantic information into a unified Mixture of Experts (MOE) framework from an information-theoretic perspective to alleviate the data-missing problem.
	
	\item To address the intrinsic conflict between consistency and complementarity, we propose the C$^2$MOE model, which decouples the two objectives via specialized experts and adaptive routing. C$^2$MOE dynamically selects the most suitable expert for each input, enabling targeted learning without mutual interference.
	
	\item Extensive experiments are conducted on multiple datasets to demonstrate that the proposed C$^2$MOE outperforms state-of-the-art methods for incomplete multimodal emotion recognition in conversation.
\end{itemize}

\section{Related Work}

\subsection{Incomplete Modality Learning}

Incomplete multimodal learning aims to address modality missingness in real-world settings, where data acquisition constraints or environmental factors often lead to partial observations \cite{zhao2021missing,duan2023alignment, li2024correlation, zhu2025proxy, lang2025retrieval, dai2025hierarchical, shou2026graph, shou2024efficient}. Existing approaches fall into two main paradigms: shared subspace learning and modality recovery. Shared subspace methods align modalities by learning a common low-dimensional representation that maximizes cross-modal correlation \cite{wang2015deep,andrew2013deep,zhang2020deep,pham2019found, shou2025spegcl, ai2026paradigm}. However, overemphasizing correlation may overlook modality-specific discriminative cues crucial for downstream tasks. In contrast, modality recovery methods explicitly reconstruct missing modalities from observed ones. These include simple heuristics, such as zero-filling \cite{zhang2020deep,deng2025multiplex} or mean imputation \cite{andrew2013deep}, as well as deep generative strategies that leverage powerful representational models. Notably, cycle-consistent cross-modal reconstruction \cite{pham2019found,zhao2021missing}, cascaded residual autoencoders \cite{tran2017missing}, and graph-based approaches \cite{lian2023gcnet} have shown promise in capturing complex inter-modal dependencies for effective recovery.

\begin{figure*}
	\centering
	\includegraphics[width=1\linewidth]{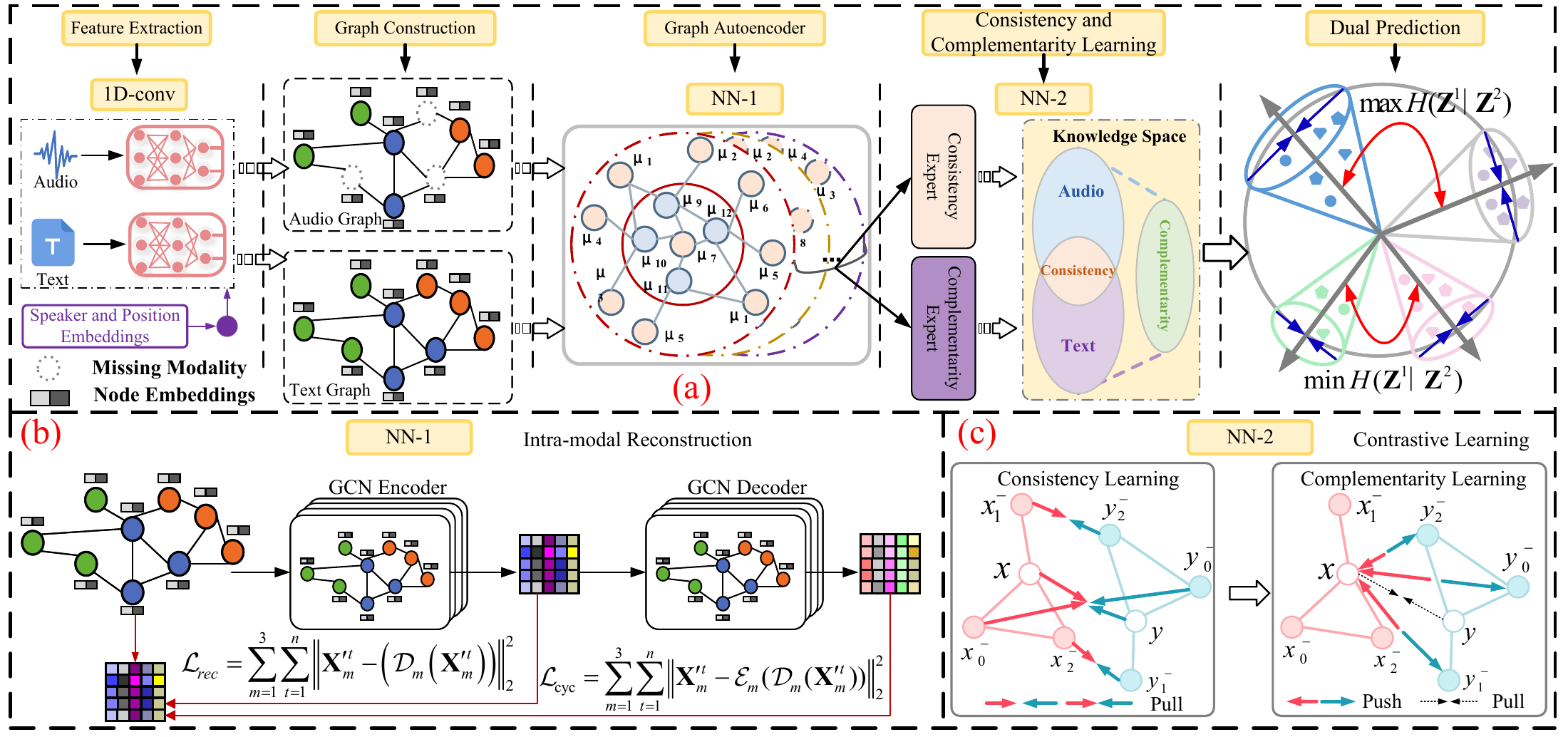}
	\caption{The overall architecture of the proposed C$^2$MOE framework. The model first extracts multimodal features using 1D convolution together with speaker and positional embeddings. NN-1 performs intra-modal reconstruction to retain modality-specific information for stable recovery. NN-2 introduces a consistency expert and a complementarity expert within an information-theoretic space, enabling multimodal knowledge to be disentangled into shared and modality-unique components. The consistency expert enhances cross-modal predictability, while the complementarity expert maximizes conditional entropy to preserve modality-specific cues. Based on these two experts, C$^2$MOE forms a dual-branch prediction mechanism for robust imputation under missing modalities (b) NN-1: intra-modal reconstruction. (c) NN-2: consistency and complementarity contrastive learning.}
	\vspace{-3mm}
	\label{fig:archi}
\end{figure*}

\subsection{Contrastive Learning}
As an efficient unsupervised learning model, contrastive learning (CL) has achieved remarkable results in feature learning \cite{lin2021completer, shou2025contrastive, shou2025revisiting, shou2025dynamic, shou2025gsdnet}. The core idea of CL is to construct a feature space in which similar data are mapped to close positions to each other, while different data points are pushed to far positions \cite{chen2020simple}, \cite{deng2018triplet}, \cite{hadsell2006dimensionality}, \cite{li2021contrastive}, \cite{tian2020contrastive}. Recent research has revealed a key factor behind the success of contrastive learning: maximization of mutual information. For example, MoCo \cite{he2020momentum} optimizes the contrastive learning loss function to maximize the similarity between the query feature and the corresponding key feature while minimizing the similarity between the query feature and other irrelevant key features. CPC  \cite{oord2018representation} attempts to maximize the similarity between the true future information and the current context while minimizing the similarity between the current context and negative samples. Both MoCo and CPC maximize mutual information by minimizing the InfoNCE loss, which can be regarded as a lower bound for maximizing mutual information.

\section{Problem Definition}

In multimodal learning, we denote an example with $M$ heterogeneous modalities as $(X^{1}, X^{2}, \ldots, X^{M})$, where $X^{m}$ corresponds to the input of the $m$-th modality. In real-world applications, some modalities may be absent due to sensor noise, transmission errors, or behavioral variations during data collection. To formally describe this incomplete setting, we introduce an observability indicator $\lambda^{m} \in \{0,1\}$, where $\lambda^{m}=1$ indicates that modality $m$ is available and $\lambda^{m}=0$ indicates it is missing. This yields the observed modality set $I^{\mathrm{obs}}=\{m \mid \lambda^{m}=1\}$ and the missing modality set $I^{\mathrm{miss}}=\{m \mid \lambda^{m}=0\}$. Our objective is to leverage the information contained in $I^{\mathrm{obs}}$ to infer and reconstruct the representations of modalities in $I^{\mathrm{miss}}$, thereby mitigating the adverse effects of data incompleteness and enabling the construction of more coherent, informative, and fusion-ready multimodal representations for downstream tasks.

\section{Proposed Method}

\subsection{Modality Encoder}
\label{sec:4.2}

Since the raw features of text, audio, and video differ substantially in dimension and scale, directly using them for missing modality recovery leads to distribution mismatch and hinders semantic alignment. To mitigate this issue and map all unimodal representations into a unified feature space, we employ a 1D-CNN layer \cite{10109845} to perform cross-modal feature projection and alignment as follows:
\begin{equation}
	\small
	\begin{aligned}
		\mathbf{X}^{\mathrm{m}^\prime}=\mathrm{Conv}1\mathbf{D}\left(\mathbf{X}^{\mathrm{m}},l^{m}\right)\in\mathbb{R}^{N\times d},\,m\in\{t,a,v\}
\end{aligned}\end{equation}
where $l^m$ represents the size of the 1D-Conv Kernel corresponding to the $m$-th modality, $N$ represents the total number of sentences in the entire dialogue sequence, and $d$ represents the dimension of the shared latent space.

To enable awareness of utterance order and strengthen contextual modeling, we add positional embeddings to the convolution-processed sequence as follows:
\begin{equation}
	\small
	\begin{aligned}
		{\bf PE}_{(pos,2i)}=\sin\left(\frac{pos}{10000^{2i/d}}\right),
		{\bf PE}_{(pos,2i+1)}=\cos\left(\frac{pos}{10000^{2i/d}}\right)
\end{aligned}\end{equation}
where $pos$ represents the index of the sequence, and dimension $i$ represents the index of the feature dimension.

We add positional embeddings to the convolutional sequence to enhance the model’s ability to capture temporal order and positional dependencies as follows:
\begin{equation}
	\small
	\mathbf{X}^{\mathrm{m}^{\prime}}=\mathbf{X}^{\mathrm{m}^{\prime}} + {\bf PE}
\end{equation}

\subsection{Intra-modal Reconstruction}
\label{sec:4.3}

Traditional multimodal learning relies on the synergy of multiple modalities, but when some modalities are missing, the learning ability of a single modality becomes particularly important. Intra-modal reconstruction enables each modality to independently learn effective feature representations, ensuring that the performance of the model will not be significantly reduced even if other modalities are missing. Specifically, for each modality, we use a graph autoencoder (GAE) \cite{kipf2016variational} to learn its latent representation $\mathbf{Z^m}$ to effectively capture the structural information and feature distribution within the modality. We first minimize the difference between the original data $\mathbf{X}^{m'}$ and the reconstructed data $\mathbf{Z^m}$ to ensure that the learned latent representation can retain the core information of the modality as follows:
\begin{equation}
	\mathcal{L}_{rec}=\sum_{m=1}^3\sum_{t=1}^n\left\|\mathbf{X}^{m'}_{t}-\left(\mathcal{D}^m\left(\mathbf{X}^{m'}_{t}\right)\right)\right\|_2^2
\end{equation}
where $\mathbf{X}^{m'}_{t}$ is the $t$-th sample of $\mathbf{X}^{m'}$, $\mathcal{D}_m$ is the graph encoder for the $m$-th modality. To ensure the semantic consistency between the encoded latent features $Z_m$ and the original data $\mathbf{X'}_m$, we introduce the consistency loss as follows:
\begin{equation}
	\mathcal{L}_{\mathrm{cyc}}=\sum_{m=1}^3\sum_{t=1}^n\left\|\mathbf{X}^{m'}_{t}-\mathcal{E}^{m}(\mathcal{D}^{m}(\mathbf{X}^{m'}_{t}))\right\|_{2}^{2}
\end{equation}
where $\mathcal{E}_{m}$ is the graph decoder for the $m$-th modality. Therefore, the latent representation of $t$-th sample in $m$-th modality is as follows:
\begin{equation}
	\mathbf{Z^m} = \mathcal{D}^m\left(\mathbf{X}^{m'}_{t}\right)
\end{equation}

\subsection{Consistency and Complementary-Guided MoE}
\label{sec:4.4}

To enhance the robustness and generalization of multimodal representation learning, we propose a consistency and complementary-guided Mixture-of-Experts (C$^2$MoE) framework that explicitly decomposes modality interactions into consistent and complementary components. This framework incorporates two specialized experts, where the consistency contrastive expert is responsible for capturing semantically aligned and modality-invariant patterns, while the complementary contrastive expert is designed to explore diverse and modality-specific signals. A shared routing network dynamically allocates each input pair to the appropriate expert by predicting soft gating weights, enabling flexible and data-adaptive expert collaboration. In the following, we discuss the detail of the C$^2$MoE framework.

\textbf{Information-Theoretic Foundations of MoE.} From an information-theoretic perspective, multimodal learning requires capturing both cross-modal consistency and modality-specific complementarity. To model consistency, we maximize the mutual information between modalities, encouraging their shared and predictable components to align. To model complementarity, we maximize the conditional entropy, which preserves modality-unique signals that cannot be inferred from other modalities. Formally, this corresponds to maximizing:
\begin{equation}
	\max_{\mathcal{H}_c} I(\mathbf{Z}^i; \mathbf{Z}^j) \quad \text{and} \quad \max_{\mathcal{H}_s} H(\mathbf{Z}^i \mid \mathbf{Z}^j)
\end{equation}
where $\mathcal{H}_c$ and $\mathcal{H}_s$ are the consistency and complementarity experts, respectively. Crucially, we implement these objectives through a dual-expert MoE architecture:
\begin{equation}
	\mathbf{Z}^m = \sum_{k=1}^{K} \alpha_k^m \cdot \mathcal{H}_k(\mathbf{Z}^m)
\end{equation}
where $\mathcal{H}_k$ are the $K$-th experts, and $\alpha_k^m$ are the routing weights. The key innovation is that each expert is specialized for either consistency or complementarity, and the routing mechanism dynamically selects the appropriate experts.

\textbf{Consistency Contrastive Expert.} To explicitly model the shared semantic patterns across modalities, we design a consistency contrastive expert. This expert specializes in capturing cross-modal 
consistency by maximizing the statistical dependency between modality representations. Given the latent representations $(\mathbf{Z}_i^1, \mathbf{Z}_i^2)$ of two modalities for 
sample $i$, the routing network produces a soft assignment vector 
$\mathbf{g}_i = [g_i^{C}, g_i^{R}]$, where $g_i^{C}$ is the gating weight for the 
consistency expert and $g_i^{R}$ for the complementary expert. In this branch, both 
modalities are passed through a shared expert function $f_C(\cdot)$:

\begin{equation}
	\widetilde{\mathbf{Z}}_i^1 = f_C(\mathbf{Z}_i^1), 
	\qquad 
	\widetilde{\mathbf{Z}}_i^2 = f_C(\mathbf{Z}_i^2)
\end{equation}

To enforce the learning of consistent structures, we introduce a 
consistency contrastive loss based on mutual information (MI) maximization:

\begin{equation}
	\mathcal{L}_{\mathrm{cl}}
	=
	-
	\sum_{i=1}^{n}
	I(\widetilde{\mathbf{Z}}_i^1; \widetilde{\mathbf{Z}}_i^2)
	+
	\alpha \left(
	H(\widetilde{\mathbf{Z}}_i^1)
	+
	H(\widetilde{\mathbf{Z}}_i^2)
	\right)
\end{equation}
where $I(\cdot;\cdot)$ denotes mutual information and $H(\cdot)$ is the entropy regularization 
term that prevents expert collapse. To compute mutual information, we estimate the joint 
distribution of discretized or soft-assignment features from the two modalities \cite{lin2022dual}. Specifically, 
a co-activation matrix $\mathbf{P} \in \mathbb{R}^{D \times D}$ is constructed as:
\begin{equation}
	\mathbf{P}
	=
	\frac{1}{m}
	\sum_{i=1}^m
	\widetilde{\mathbf{Z}}_i^1
	\left(\widetilde{\mathbf{Z}}_i^2\right)^{\top}
	\tag{10}
\end{equation}
where $D$ denotes the number of discretization bins or latent dimensions used to estimate 
feature distributions. The marginal distributions are obtained by summing rows and columns 
of $\mathbf{P}$. The final discrete MI objective is written as:

\begin{equation}
	\mathcal{L}_{\mathrm{cl}}
	=
	-
	\sum_{d=1}^{D}
	\sum_{d'=1}^{D}
	P_{dd'}
	\log
	\left(
	\frac{P_{dd'}}{P_d \cdot P_{d'}}
	\right)
\end{equation}
which encourages to maximize statistical consistency between modalities. Through 
joint training with routing, the consistency expert specializes in capturing cross-modal 
shared structures, while the complementary expert focuses on modality-specific information.

\textbf{Consistency Expert Prediction.} To infer modality-consistent representations within the proposed dual-expert MoE framework, we design a consistency expert prediction mechanism that estimates the missing or incomplete modality from its consistent counterpart. This allows the 
consistency contrastive expert to perform semantic reconstruction through 
conditional generative modeling. Formally, given paired modality representations 
$(\mathbf{Z}_i^1, \mathbf{Z}_i^2)$, the goal is to minimize the conditional entropy 
$H(\mathbf{Z}_i^1 \mid \mathbf{Z}_i^2)$ so that $\mathbf{Z}_i^1$ can be deterministically 
inferred from $\mathbf{Z}_i^2$:
\begin{equation}
	H(\mathbf{Z}_i^1 \mid \mathbf{Z}_i^2)
	=
	-
	\mathbb{E}_{p(\mathbf{Z}_i^1, \mathbf{Z}_i^2)}
	\left[
	\log p(\mathbf{Z}_i^1 \mid \mathbf{Z}_i^2)
	\right]
	\tag{12}
\end{equation}

Since directly computing the true posterior 
$p(\mathbf{Z}_i^1 \mid \mathbf{Z}_i^2)$ is intractable, we introduce a 
variational approximation $Q(\mathbf{Z}_i^1 \mid \mathbf{Z}_i^2)$, parameterized by a 
neural network. Maximizing the conditional likelihood leads to the following 
variational lower bound:

\begin{equation}
	\begin{aligned}
		&\mathbb{E}_{p(\mathbf{Z}_i^1, \mathbf{Z}_i^2)}
		\left[
		\log p(\mathbf{Z}_i^1 \mid \mathbf{Z}_i^2)
		\right] \\
		&\geq 
		\mathbb{E}_{p(\mathbf{Z}_i^1, \mathbf{Z}_i^2)}
		\left[
		\log Q(\mathbf{Z}_i^1 \mid \mathbf{Z}_i^2)
		\right] \\
		&-
		\mathcal{D}_{\mathrm{KL}}
		\left(
		Q(\mathbf{Z}_i^1 \mid \mathbf{Z}_i^2)
		\parallel
		p(\mathbf{Z}_i^1 \mid \mathbf{Z}_i^2)
		\right)
	\end{aligned}
\end{equation}

For practical optimization, we assume that the variational distribution is Gaussian:

\[
Q(\mathbf{Z}_i^1 \mid \mathbf{Z}_i^2)
=
\mathcal{N}
\left(
G^{(C)}(\mathbf{Z}_i^2),
\;
\sigma^2 I
\right)
\]
where $G^{(C)}(\cdot)$ denotes the consistency prediction network and $\sigma^2 I$ 
represents prediction uncertainty. Under the Gaussian assumption, maximizing the lower 
bound becomes equivalent to minimizing the reconstruction loss:

\begin{equation}
	\mathcal{L}_{\mathrm{cl}}'
	=
	\mathbb{E}_{p(\mathbf{Z}_i^1, \mathbf{Z}_i^2)}
	\left\|
	\mathbf{Z}_i^1
	-
	G^{(C)}(\mathbf{Z}_i^2)
	\right\|_2^2
	\tag{14}
\end{equation}

Once training is complete, it can be used to directly infer the missing modality 
representation:
\begin{equation}
	\widetilde{\mathbf{Z}}_i^1
	=
	G^{(C)}(\mathbf{Z}_i^2)
	\tag{15}
\end{equation}

\textbf{Complementary Contrastive Expert.} While the consistency expert focuses on modality-invariant shared semantics, 
real-world multimodal data often exhibit rich complementary information that is 
specific to each modality. To model such modality-specific diversity, we introduce the 
complementary contrastive expert, which is designed to 
capture non-overlapping but semantically coordinated representations across modalities. Let $\mathbf{Z}_i^1$ and $\mathbf{Z}_i^2$ denote the representations from two modalities 
for sample $i$. These features are passed through a shared expert function 
$f_R(\cdot)$, yielding complementary representations:

\begin{equation}
	\widehat{\mathbf{Z}}_i^1 = f_R(\mathbf{Z}_i^1), 
	\quad 
	\widehat{\mathbf{Z}}_i^2 = f_R(\mathbf{Z}_i^2)
\end{equation}

To enforce cross-modal complementarity, we design a loss that encourages each modality to retain rich information while reducing redundancy by maximizing entropy and penalizing mutual information as follows:

\begin{equation}
	\mathcal{L}_{\mathrm{p}} 
	= 
	- \sum_{i=1}^{n} \left[
	H(\widehat{\mathbf{Z}}_i^1) 
	+ 
	H(\widehat{\mathbf{Z}}_i^2)
	\right]
	+ 
	\beta \cdot I(\widehat{\mathbf{Z}}_i^1; \widehat{\mathbf{Z}}_i^2)
\end{equation}
where $\beta > 0$ is a hyperparameter balancing the two terms. To estimate mutual information in discrete space, we compute a joint co-activation matrix:
\begin{equation}
	\mathbf{P}^{\mathrm{comp}} 
	= 
	\frac{1}{m} 
	\sum_{i=1}^{m} 
	\widehat{\mathbf{Z}}_i^1 
	\left( 
	\widehat{\mathbf{Z}}_i^2 
	\right)^{\top}
\end{equation}
with marginals $P_d^{(1)}$ and $P_{d'}^{(2)}$ computed by row and column summation, respectively. 
The entropy and mutual information terms can then be estimated as:

\begin{equation}
	\begin{aligned}
		\mathcal{L}_{\mathrm{p}} 
		= 
		&- 
		\sum_{d=1}^{D} 
		\left[ 
		H\left( P_d^{(1)} \right) 
		+ 
		H\left( P_d^{(2)} \right) 
		\right] \\
		&+ 
		\beta 
		\sum_{d=1}^{D} 
		\sum_{d'=1}^{D} 
		P^{\mathrm{comp}}_{dd'} 
		\log 
		\left( 
		\frac{P^{\mathrm{p}}_{dd'}}{P_d^{(1)} \cdot P_{d'}^{(2)}}
		\right)
	\end{aligned}
\end{equation}

\textbf{Complementary Expert Prediction.} To reconstruct missing modality-specific information that is semantically relevant yet non-overlapping, we introduce a complementary expert prediction mechanism. Unlike consistency prediction 
which targets shared semantics, this module focuses on capturing diverse signals unique to 
each modality. In contrast to consistent expert predictions, its training objective is simplified to maximizing the expected reconstruction error between the true value and the predicted complementary features.
\begin{equation}
	\mathcal{L}_{\mathrm{p}}' = 
	- 
	\mathbb{E}_{p(\tilde{\mathbf{Z}}_i, \mathbf{Z}_i^j)}
	\left\|
	\tilde{\mathbf{Z}}_i - \mathcal{E}^{(j)}(\mathbf{Z}_i^j)
	\right\|_2^2
\end{equation}

Once training is complete, the complementary feature can be estimated via:
\begin{equation}
	\tilde{\mathbf{Z}}_i = \mathcal{E}^{(j)}(\mathbf{Z}_i^j) = 
	\mathcal{E}^{(j)}(\mathcal{D}^{(j)}(X^j))
\end{equation}
where $\mathcal{D}^{(j)}$ is the encoder for modality $j$.

\begin{table*}[htbp]
	\centering
	\caption{The performance of different methods is shown under different missing modalities on the CMU-MOSI and CMU-MOSEI datasets. The values reported in each cell represent the $\mathrm{ACC_2}$/$\mathrm{F1}$/$\mathrm{ACC_7}$. Bold indicates the best performance.}
	\label{TABLE:1}
	\vspace{-2mm}
	\resizebox{1.0\textwidth}{!}{
		\begin{tabular}{l|c|ccccccc}
			\toprule
			Datasets &Available & MCTN  & MMIN  & GCNet & DiCMoR & IMDer & GSDNet & C$^2$MOE (Ours) \\
			\midrule
			\multirow{8}{*}{CMU-MOSI} 
			& $\{l\}$  & 79.1/79.2/41.0 & 83.8/83.8/41.6 & 83.7/83.6/42.3 & 84.5/84.4/44.3 & {84.8}/{84.7}/{44.8} &   86.4/86.6/45.7     & \textbf{87.3/87.7/48.0} \\
			& $\{v\}$  & 55.0/54.4/16.3 & 57.0/54.0/15.5 & 56.1/55.7/16.9 & 62.2/60.2/20.9 & {61.3}/{60.8}/{22.2} &  64.1/63.7/25.3     & \textbf{65.3/64.5/26.1}\\
			& $\{a\}$  & 56.1/54.5/16.5 & 55.3/51.5/15.5 & 56.1/54.5/16.6 &62.2/60.2/20.9 & {62.0}/{62.2}/{22.0} &  64.4/64.1/24.6     & \textbf{65.6/65.5/26.4}\\
			& $\{l,v\}$  & 81.1/81.2/42.1 & 83.8/83.9/42.0 & 84.3/84.2/43.4 &85.5/85.4/45.2 & {85.5}/{85.4}/{45.3} &   86.5/86.4/46.7    & \textbf{87.6/87.9/47.4}\\
			& $\{l,a\}$  & 81.0/81.0/43.2 & 84.0/84.0/42.3 & 84.5/84.4/43.4 &85.5/85.5/44.6 & {85.4}/{85.3}/{45.0} &  86.7/86.6/46.8     & \textbf{88.0/87.8/47.6}\\
			& $\{v,a\}$  & 57.5/57.4/16.8 & 60.4/58.5/19.5 & 62.0/61.9/17.2 &64.0/63.5/21.9 & {63.6}/{63.4}/{23.8} &   65.2/64.8/24.9    & \textbf{66.7/65.5/25.2}\\
			& $\{l,v,a\}$  & 81.4/81.5/43.4 & 84.6/84.4/44.8 & 85.2/85.1/44.9  & 85.7/85.6/45.3 & {85.7}/{85.6}/{45.3} &   87.7/87.3/46.8    & \textbf{88.5/88.8/47.4}\\
			& Average & 70.2/69.9/31.3 & 72.7/71.4/31.6 & 73.1/72.8/32.1 & 75.4/75.1/34.7 &  {75.5}/{75.3}/{35.5} &  77.3/77.1/37.3     & \textbf{78.4/78.2/38.3}\\
			\midrule
			\midrule
			\multirow{8}{*}{CMU-MOSEI} 
			& $\{l\}$ & 82.6/82.8/50.2 & 82.3/82.4/51.4 & 83.0/83.2/51.2 & 84.2/84.3/52.4 & {84.5}/{84.5}/{52.5} &   86.6/86.1/55.3    & \textbf{87.3/87.1/56.7}\\
			& $\{v\}$ & 62.6/57.1/41.6 & 59.3/60.0/40.7 & 61.9/61.6/41.7 & 63.6/63.6/42.0 & {63.9}/{63.6}/{42.6} &   65.1/65.7/44.9    & \textbf{66.8/66.6/45.7}\\
			& $\{a\}$ & 62.7/54.5/41.4 & 58.9/59.5/40.4 & 60.2/60.3/41.1 & 62.9/60.4/41.4 & {63.8}/{60.6}/{41.7} &   64.6/64.2/43.1    & \textbf{66.4/65.5/44.4}\\
			& $\{l,v\}$ & 83.2/83.2/50.4 & 83.8/83.4/51.2 & 84.3/84.4/51.1 & 84.9/84.9/53.0 & {85.0}/{85.0}/{53.1} &   87.3/87.0/56.2    & \textbf{88.8/88.3/57.1}\\
			& $\{l,a\}$ & 83.5/83.3/50.7 & 83.7/83.3/52.0 & 84.3/84.4/51.3 & 85.0/84.9/52.7 & {85.1}/{85.1}/{53.1} &   86.2/86.4/55.5    & \textbf{87.4/87.7/56.2}\\
			& $\{v,a\}$ & 63.7/62.7/42.1 & 63.5/61.9/41.8 & 64.1/57.2/42.0 & 65.2/64.4/42.4 & {64.9}/{63.5}/{42.8} &   66.7/66.3/45.2    & \textbf{67.9/66.7/46.0}\\
			& $\{l,v,a\}$ & 84.2/84.2/51.2 & 84.3/84.2/52.4 & 85.2/85.1/51.5 & 85.1/85.1/53.4 & {85.1}/{85.1}/{53.4} &   87.3/87.2/54.9    & \textbf{88.6/88.1/56.5}\\
			& Average & 74.6/72.5/46.8 & 73.7/73.5/47.1 & 74.7/73.7/47.1 & 75.8/75.4/48.2 & {76.0}/{75.3}/{48.5} &    77.7/77.6/50.7   & \textbf{79.0/78.6/51.8}\\
			\bottomrule
	\end{tabular}}
	\vspace{-3mm}
\end{table*}

\textbf{Routing-aware Expert Integration.} To adaptively balance shared and modality-specific semantics during prediction, we introduce a routing network that dynamically fuses the outputs from 
the consistency and complementary experts based on the features of the 
observed modality representation. Given the input feature $\mathbf{Z}_i^j$ from the observed modality $j$, 
the routing network generates a soft assignment vector 
$\mathbf{g}_i = [g_i^{C}, g_i^{R}]$, where $g_i^{C}, g_i^{R} \in [0, 1]$ and 
$g_i^{C} + g_i^{R} = 1$. These weights are computed via a softmax gating mechanism as follows:
\begin{equation}
	\mathbf{g}_i = \text{softmax}(\mathbf{W} \cdot \mathbf{Z}_i^j + \mathbf{b})
\end{equation}
where $\mathbf{W}$ and $\mathbf{b}$ are learnable parameters. 
The weights $g_i^C$ and $g_i^R$ denote the importance of the consistency and 
complementary predictions, respectively. The final predicted representation is given by the weighted fusion as follows:
\begin{equation}
	\mathbf{Z}_i^{\text{pred}} = 
	g_i^{C} \cdot \bar{\mathbf{Z}}_i + 
	g_i^{R} \cdot \tilde{\mathbf{Z}}_i
\end{equation}
where $\bar{\mathbf{Z}}_i = G^{(C)}(\mathbf{Z}_i^j)$ is the consistency prediction, and 
$\tilde{\mathbf{Z}}_i = \mathcal{E}^{(j)}(\mathbf{Z}_i^j)$ is the complementary 
prediction.

\subsection{Objective Function}
To jointly optimize consistency and complementarity within the dual-expert framework, 
we define a composite objective that integrates contrastive and predictive components 
from both experts. The total loss is formulated as:
\begin{equation}
	\mathcal{L}_{\text{total}} = 
	\lambda_1 \left(\mathcal{L}_{\text{cl}} + 
	\cdot \mathcal{L}_{\text{cl}}'\right) + 
	\lambda_2 \left (\mathcal{L}_{\text{p}} + 
	\cdot \mathcal{L}_{\text{p}}'\right) + \mathcal{L}_{\text{rec}} + \mathcal{L}_{\text{cyc}}
\end{equation}
where $\lambda_1, \lambda_2$ are hyperparameters used to balance the relative importance of the four objectives.

\section{Experiments}

\subsection{Corpus Description}

To verify the effectiveness of C$^2$MOE in different conversation scenarios, we conduct extensive experiments on two benchmark datasets, including CMU-MOSI \cite{zadeh2016multimodal}, and CMU-MOSEI \cite{zadeh2018multimodal}. CMU-MOSI dataset contains video comments from social media platforms, covering a wide range of topics and emotional expressions. Each video clip contains audio data, video data, and text data. CMU-MOSEI dataset contains a large number of video comments from YouTube. The CMU-MOSI and CMU-MOSEI dataset is annotated with sentiment intensity annotations ranging from -3 (very negative) to +3 (very positive), and 0 represents neutral sentiment.  Following previous work \cite{liang2021attention, lv2021progressive}, we evaluate the MERC performance using the following metrics: 7-class accuracy ($\mathrm{ACC_7}$), binary accuracy ($\mathrm{ACC_2}$), and $\mathrm{F1}$ score.

\subsection{Implementation Details}

We evaluate model performance under varying degrees of modality incompleteness. The missing rate $\eta$ is defined as $\eta = 1 - \frac{\sum_{i=1}^{L} m_i}{L \times M}$
where $m_i$ denotes the number of available modalities for the $i$-th sample, $L$ is the number of samples, and $M$ is the total number of modalities. For each sample, modalities are randomly masked according to $\eta$, while ensuring that at least one modality remains ($1 \leq m_i \leq M$), which bounds $\eta \leq \frac{M-1}{M}$. For $M = 3$, we vary $\eta$ from $0.0$ to $0.7$ in steps of $0.1$, following prior work~\cite{lian2023gcnet}. The same missing rate is applied across training, validation, and testing for fair comparison.

For each utterance, we extract acoustic, lexical, and visual features using pre-trained models. Specifically, \textit{wav2vec-large}~\cite{schneider2019wav2vec} is used to extract 512-dimensional acoustic features, MA-Net~\cite{zhao2021learning} to extract 1024-dimensional facial features, and DeBERTa~\cite{hedeberta} to extract 1024-dimensional lexical features. All experiments are implemented in PyTorch on an NVIDIA RTX 4090 GPU with a batch size of 16. We adopt early stopping and train the model for up to 60 epochs until convergence.


\subsection{Baselines}
To evaluate the performance of our proposed C$^2$MOE, we compare it to the state-of-the-art incomplete multimodal learning methods, including MCTN \cite{pham2019found}, MMIN \cite{zhao2021missing}, GCNet \cite{lian2023gcnet}, DiCMoR \cite{wang2023distribution}, IMDer \cite{wang2023incomplete}, and GSDNet \cite{shou2025gsdnet}.

\begin{table*}[ht]
	\centering
	\caption{The performance of different methods is shown at different missing ratios on the CMU-MOSI and CMU-MOSEI datasets. The values reported in each cell represent the $\mathrm{ACC_2}$/$\mathrm{F1}$/$\mathrm{ACC_7}$. Bold indicates the best performance.}
	\label{TABLE:2}
	\vspace{-2mm}
	\resizebox{1.0\textwidth}{!}{
		\begin{tabular}{l|c|ccccccc}
			\toprule
			Datasets & Missing Rate & MCTN & MMIN & GCNet & DiCMoR & IMDer &  GSDNet & C$^2$MOE (Ours) \\
			\midrule
			\multirow{9}{*}{CMU-MOSI} 
			& 0.0  & 81.4/81.5/43.4 & 84.6/84.4/44.8 & 85.2/85.1/44.9 & 85.7/85.6/45.3 & 85.7/85.6/45.3 & 87.7/87.3/46.8     & \textbf{88.5/88.8/47.4}\\
			& 0.1  & 78.4/78.5/39.8 & 81.8/81.8/41.2 & 82.3/82.3/42.1 & 83.9/83.9/43.6 & 84.9/84.8/44.8 &  87.1/86.5/46.2    & \textbf{88.4/88.0/47.8}\\
			& 0.2  & 75.6/75.7/38.5 & 79.0/79.1/38.9 & 79.4/79.5/40.0 & 83.9/83.9/43.6 & 83.5/83.4/44.3 &  86.4/86.1/45.2    & \textbf{87.1/87.5/46.5}\\
			& 0.3  & 71.3/71.2/35.5 & 76.1/76.2/36.9 & 77.2/77.2/38.2 & 80.4/80.2/40.6 & 81.2/81.0/42.5 &  85.2/85.0/44.3    & \textbf{86.7/86.1/45.8}\\
			& 0.4  & 68.0/67.6/32.9 & 71.7/71.6/34.9 & 74.3/74.4/36.6 & 77.9/77.7/37.6 & 78.6/78.5/39.7 &   83.3/82.9/42.1   & \textbf{85.2/84.7/43.7}\\
			& 0.5  & 65.4/64.8/31.2 & 67.2/66.5/32.2 & 70.0/69.8/33.9 & 76.7/76.4/36.4 & 76.2/75.9/37.9 &  81.2/81.1/40.6    & \textbf{82.4/82.4/41.5}\\
			& 0.6  & 63.8/62.5/29.7 & 64.9/64.0/29.1 & 67.7/66.7/29.8 & 73.3/73.0/32.7 & 74.7/74.0/35.8 &   80.1/79.7/38.7   & \textbf{81.7/81.3/39.2}\\
			& 0.7  & 61.2/59.0/27.5 & 62.8/61.0/28.4 & 65.7/65.4/28.1 & 71.1/70.8/30.0 & 71.9/71.2/33.4 &  77.6/77.3/35.6    & \textbf{78.3/78.5/36.3}\\
			& Average  & 70.6/70.1/34.8 & 73.5/73.1/35.8 & 75.2/75.1/36.7 & 78.9/78.7/38.5 & 79.6/79.3/40.5 &    83.6/83.2/42.3  & \textbf{84.8/84.7/43.5}\\
			\midrule
			\midrule
			\multirow{9}{*}{CMU-MOSEI} 
			& 0.0  & 84.2/84.2/51.2 & 84.3/84.2/52.4 & 85.2/85.1/51.5 & 78.9/78.7/38.5 & 85.1/85.1/53.4 &   87.3/87.2/54.9   & \textbf{88.6/88.1/55.5} \\
			& 0.1  & 81.8/81.6/49.8 & 81.9/81.3/50.6 & 82.3/82.1/51.2 & 78.9/78.7/38.5 & 84.8/84.6/53.1 &  86.7/86.5/54.2    & \textbf{87.9/87.7/55.6}\\
			& 0.2  & 79.0/78.7/48.6 & 79.8/78.8/49.6 & 80.3/79.9/50.2 & 81.8/81.5/51.4 & 82.7/82.4/52.0 & 85.3/85.1/53.5     & \textbf{86.2/86.1/54.3}\\
			& 0.3  & 76.9/76.2/47.4 & 77.2/75.5/48.1 & 77.5/76.8/49.2 & 79.8/79.3/50.3 & 81.3/80.7/51.3 &  83.3/83.0/52.2    & \textbf{84.8/84.6/53.8}\\
			& 0.4  & 74.3/74.1/45.6 & 75.2/72.6/47.5 & 76.0/74.9/48.0 & 78.7/77.4/48.8 & 79.3/78.1/50.0 &  81.4/81.2/51.4    & \textbf{83.2/82.5/53.2}\\
			& 0.5  & 73.6/72.6/45.1 & 73.9/70.7/46.7 & 74.9/73.2/46.7 & 77.7/75.8/47.7 & 79.0/77.4/49.2 &   80.5/80.1/50.7   & \textbf{81.3/81.6/51.4}\\
			& 0.6  & 73.2/71.1/43.8 & 73.2/70.3/45.6 & 74.1/72.1/45.1 & 77.7/75.8/47.7 & 78.0/75.5/48.5 &  79.4/79.1/49.4    & \textbf{80.7/80.3/50.6}\\
			& 0.7  & 72.7/70.5/43.6 & 73.1/69.5/44.8 & 73.2/70.4/44.5 & 75.4/72.2/46.2 & 77.3/74.6/47.6 &   78.2/78.1/48.6   & \textbf{79.5/79.3/49.7}\\
			& Average  & 77.0/76.1/46.9 & 77.3/75.4/48.2 & 77.9/76.8/48.3 & 79.9/78.6/49.6 & 80.9/79.8/50.6 &  82.8/82.5/51.9    & \textbf{84.0/83.8/53.0}\\
			\bottomrule
	\end{tabular}}
	\vspace{-3mm}
\end{table*}

\section{Results and Discussion}

\subsection{Performance of Missing Modalities}

Table \ref{TABLE:1} list the quantitative results for different missing modalities and random missing rates on the CMU-MOSI and CMU-MOSEI datasets, showing the performance of each method in the case of missing modalities. C$^2$MOE achieved the best results on both datasets, verifying its superiority in dealing with the problem of missing modalities. The superior performance of C$^2$MOE may be attributed to its ability to simultaneously utilize the consistency and complementary semantic information of multimodal data, which not only helps to ensure the consistency of the restored data with the original data, but also provides additional supplementary information for the recovered data, enhancing the robustness of the model in the case of missing modalities.

\begin{table}[htbp]
	\caption{Ablation study of C$^2$MOE under average random missing ratios. The values reported in each cell represent the $\mathrm{ACC_2}$/$\mathrm{F1}$/$\mathrm{ACC_7}$. Bold indicates the best performance.}
	\label{tab:aba}
	\vspace{-2mm}
	\setlength{\tabcolsep}{0.7mm}{
		\begin{tabular}{cccccccccc}
			\toprule
			\multirow{2}{*}{$\mathcal{L}_{cl}$} & \multirow{2}{*}{$\mathcal{L}_{p}$} & \multirow{2}{*}{$\mathcal{L'}_{cl}$} & \multirow{2}{*}{$\mathcal{L'}_{p}$}   & \multicolumn{3}{c}{CMU-MOSI} & \multicolumn{3}{c}{CMU-MOSEI} \\ \cline{5-10}                   &                                        &                    &                    & $ACC_2$     & F1     & $ACC_7$     & $ACC_2$      & F1     & $ACC_7$     \\ \midrule
			$\checkmark$       & $\checkmark$              &                    &                                        &    81.4      &   81.0     &   41.2       &   81.1        &  80.8      &  50.5        \\
			&                                        & $\checkmark$       & $\checkmark$               &    78.6      &    78.2    &    39.4      &     78.3      &   77.7     &   48.2       \\
			$\checkmark$       &                                        & $\checkmark$       &                              &    74.7      &  72.0      &  33.6        &   72.1        &  71.4      &   42.3       \\
			& $\checkmark$       &                                        & $\checkmark$                          &   75.8       &   75.3     &   34.1       &    73.0       &  72.6      & 45.4         \\
			$\checkmark$       & $\checkmark$              & $\checkmark$       & $\checkmark$              &    \textbf{84.8}      &    \textbf{84.7}    &     \textbf{43.5}     &     \textbf{84.0}      &   \textbf{83.8}     &   \textbf{53.0}       \\ \bottomrule
	\end{tabular}}
	\vspace{-3mm}
\end{table}

\subsection{Performance of Missing Ratios}

Table \ref{TABLE:2} list the quantitative results for random missing rates on the CMU-MOSI and CMU-MOSEI datasets, showing the performance of each method in the case of missing rates. Compared with other MERC methods, the performance degradation of C$^2$MOE decreases with the increase of the missing modal rate, which shows that C$^2$MOE can effectively cope with the challenges brought by missing modalities and can still maintain high recognition performance under high missing rate conditions. In fact, in most restoration-based models, the performance usually experiences a significant decline as the missing modal rate increases. C$^2$MOE significantly slows down the performance degradation by better utilizing the consistency and complementarity between modalities.

\subsection{Ablation Studies}

Table~\ref{tab:aba} presents an ablation study that investigates the impact of different loss components within the C\textsuperscript{2}MOE framework. When only the consistency contrastive loss $\mathcal{L}_{\mathrm{cl}}$ is employed, the model already achieves strong baseline performance, which confirms the effectiveness of this loss in capturing modality-invariant semantic representations and enforcing semantic alignment across modalities. In contrast, using only the complementary contrastive loss $\mathcal{L}_{\mathrm{comp}}$ leads to a slight performance degradation. This suggests that emphasizing diversity alone, without sufficient consistency constraints, may introduce semantic noise and hinder stable representation learning. Furthermore, prediction-only training using $\mathcal{L}_{\mathrm{cl'}}$ and $\mathcal{L}_{\mathrm{comp'}}$ results in significantly inferior performance, indicating that reconstruction or prediction objectives without explicit semantic alignment are insufficient for robust multimodal understanding. Notably, the best performance is achieved when all four loss terms are jointly optimized. This demonstrates that contrastive and predictive objectives play complementary roles: contrastive losses enforce semantic structure, while predictive losses enhance representational completeness. These results clearly validate the effectiveness of jointly modeling both consistency and complementarity under the routing mechanism in the C\textsuperscript{2}MOE framework.

\begin{figure}[htbp]
	\centering
	\setlength{\abovecaptionskip}{0.cm}
	\subfloat[CMU-MOSI]{\includegraphics[width=0.495\linewidth]{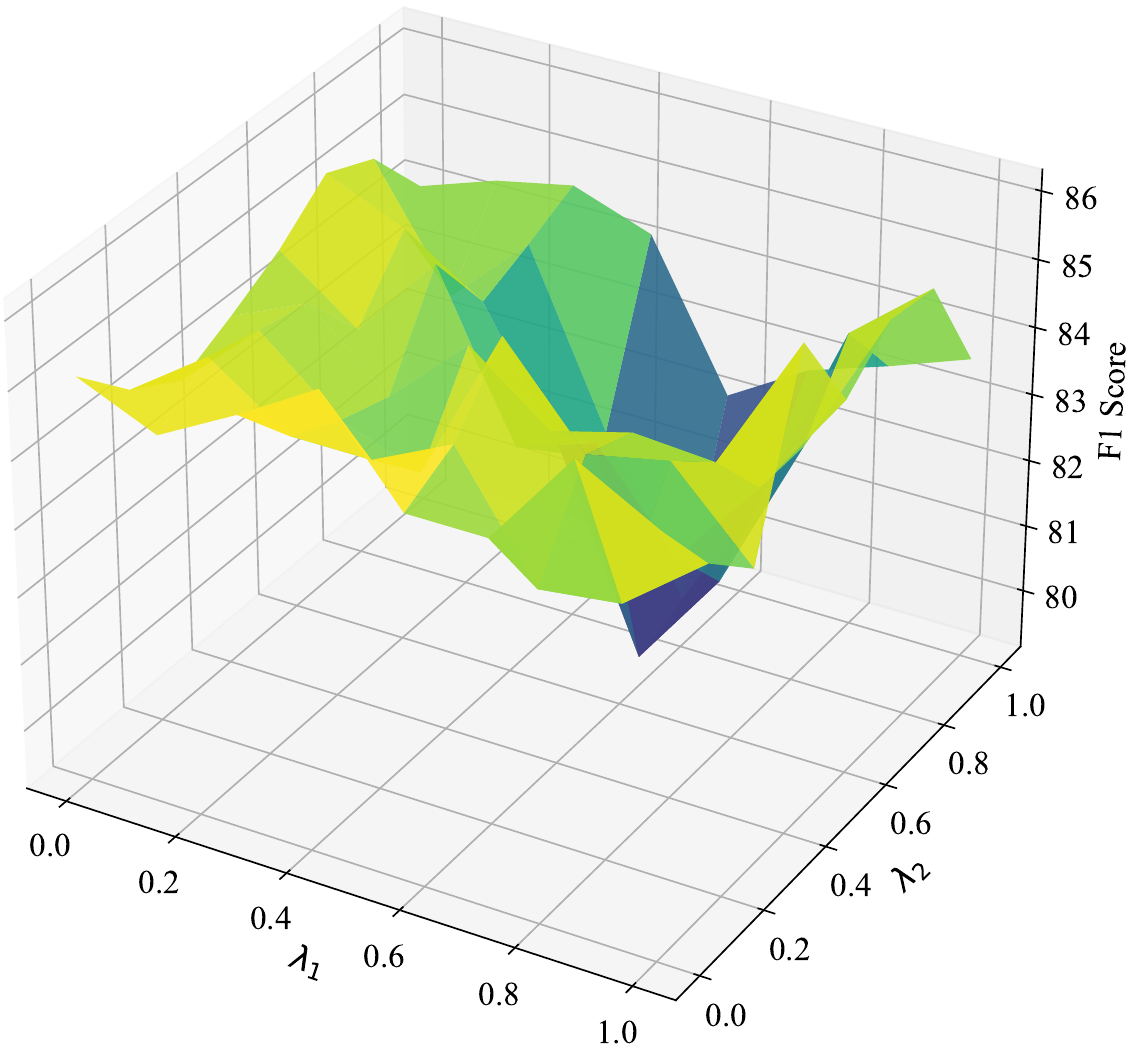}
		\label{fig:hyper_mosi}}
	\hfil
	\subfloat[CMU-MOSEI]{\includegraphics[width=0.48\linewidth]{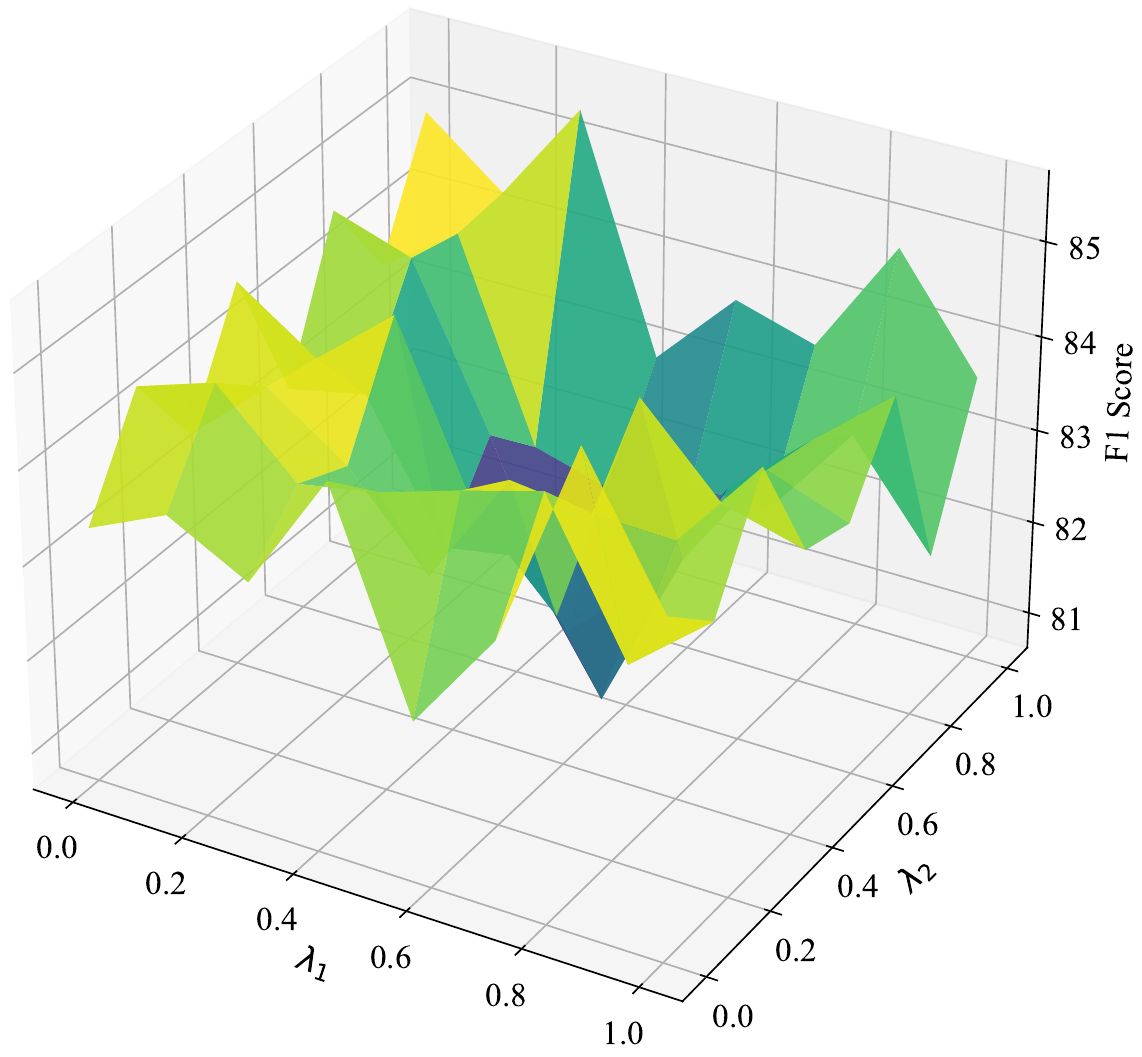}
		\label{fig:hyper_mosei}}
	\caption{F1 score performance of the proposed model under varying hyperparameters $\lambda_1$ and $\lambda_2$ on two benchmark datasets.}
	\label{fig.hyper}
	\vspace{-3mm}
\end{figure}

\begin{figure*}
	\centering
	\includegraphics[width=1\linewidth]{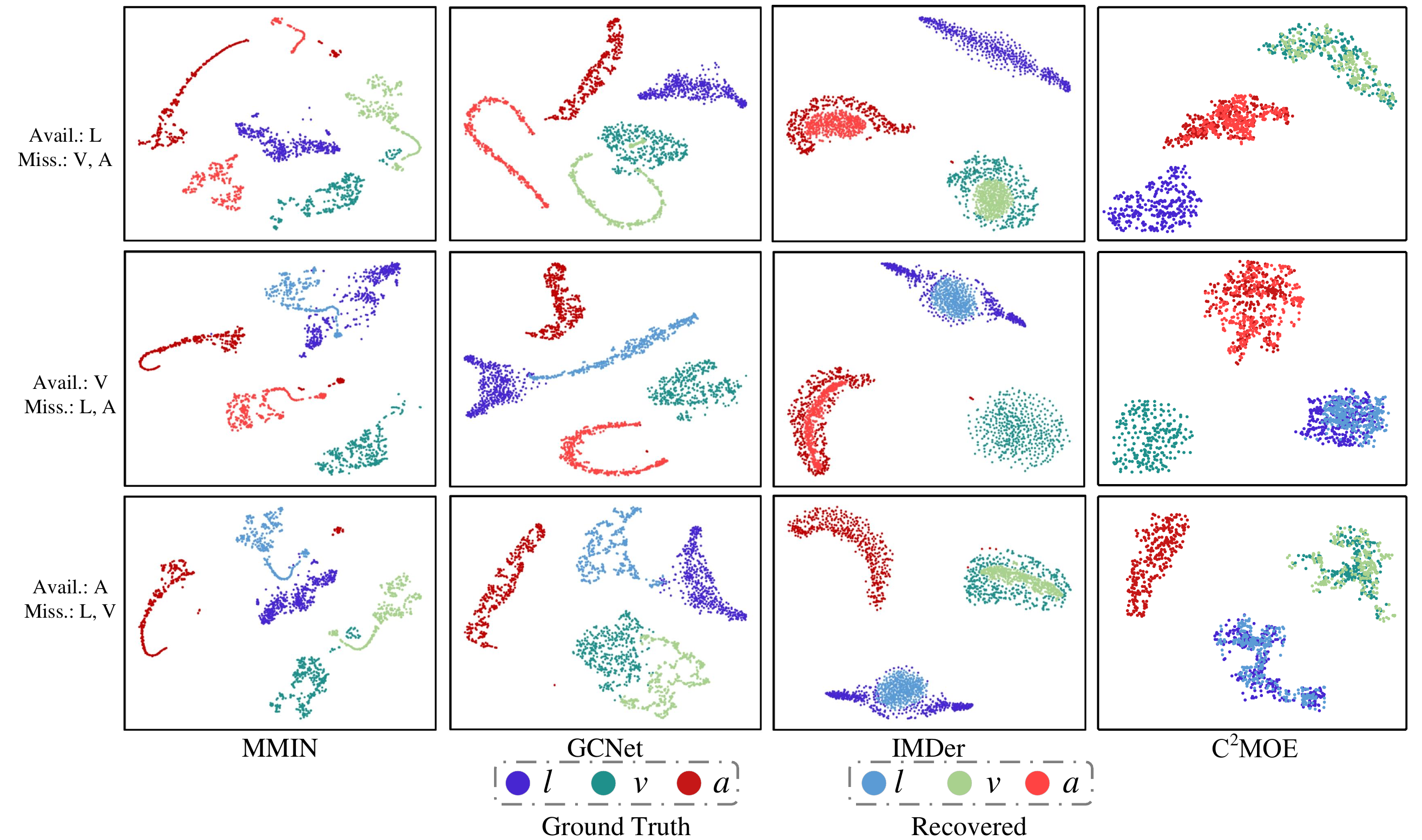}
	\caption{Visualization of restored modalities. Avail. indicates available.}
	\label{fig:enter-label}
	\vspace{-3mm}
\end{figure*}

\subsection{Hyperparameter Sensitivity Analysis}


Fig.~\ref{fig:hyper} illustrates the F1-score performance of the proposed model on the CMU-MOSI and CMU-MOSEI datasets under different settings of $\lambda_1$ and $\lambda_2$, which control the relative weights of the consistency and complementary contrastive losses, respectively. As shown in the figure, the model exhibits stable performance within a moderate range of both hyperparameters, indicating that C\textsuperscript{2}MOE is not overly sensitive to precise hyperparameter tuning. Notably, clear performance peaks are observed when $\lambda_1 \in [0.6, 0.8]$ and $\lambda_2 \in [0.3, 0.5]$, suggesting that an appropriate balance between consistency and complementarity is crucial for optimal performance. When either hyperparameter is set to an extremely low or high value, the model performance degrades noticeably. This phenomenon implies that underemphasizing consistency may weaken semantic alignment across modalities, while overemphasizing it may suppress useful complementary information. Similarly, assigning excessive weight to the complementary loss may introduce semantic noise, whereas insufficient emphasis may limit the model’s ability to leverage modality-specific cues. Overall, these results highlight the importance of jointly modeling consistency and complementarity to learn balanced and robust multimodal representations.

\begin{figure}[htbp]
	\centering
	\setlength{\abovecaptionskip}{0.cm}
	\subfloat[CMU-MOSI]{\includegraphics[width=0.495\linewidth]{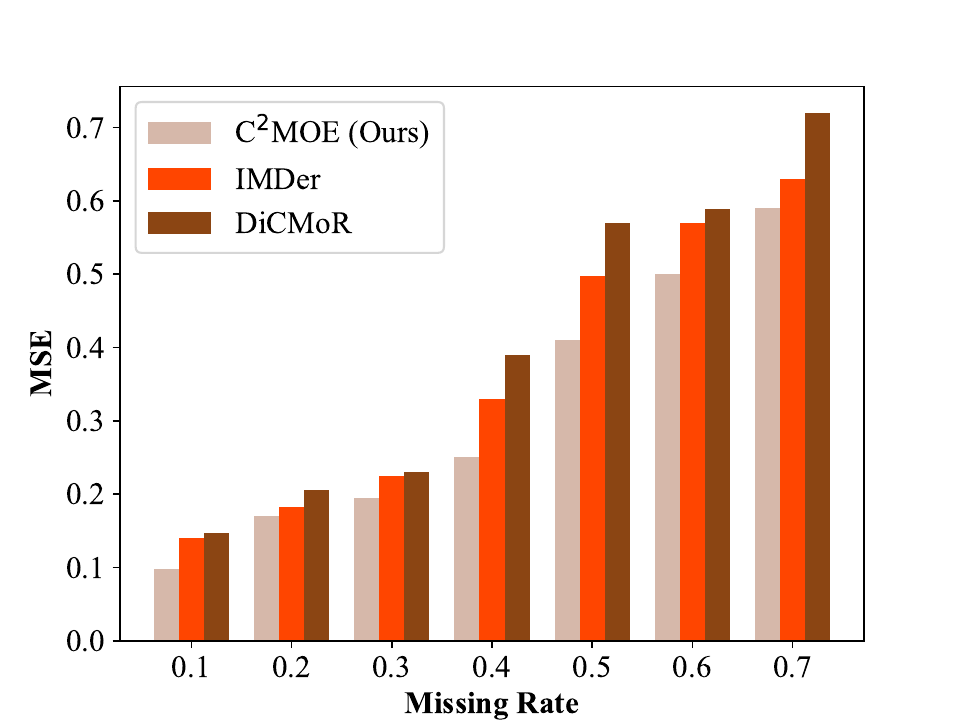}
		\label{fig:mosi}}
	\hfil
	\subfloat[CMU-MOSEI]{\includegraphics[width=0.48\linewidth]{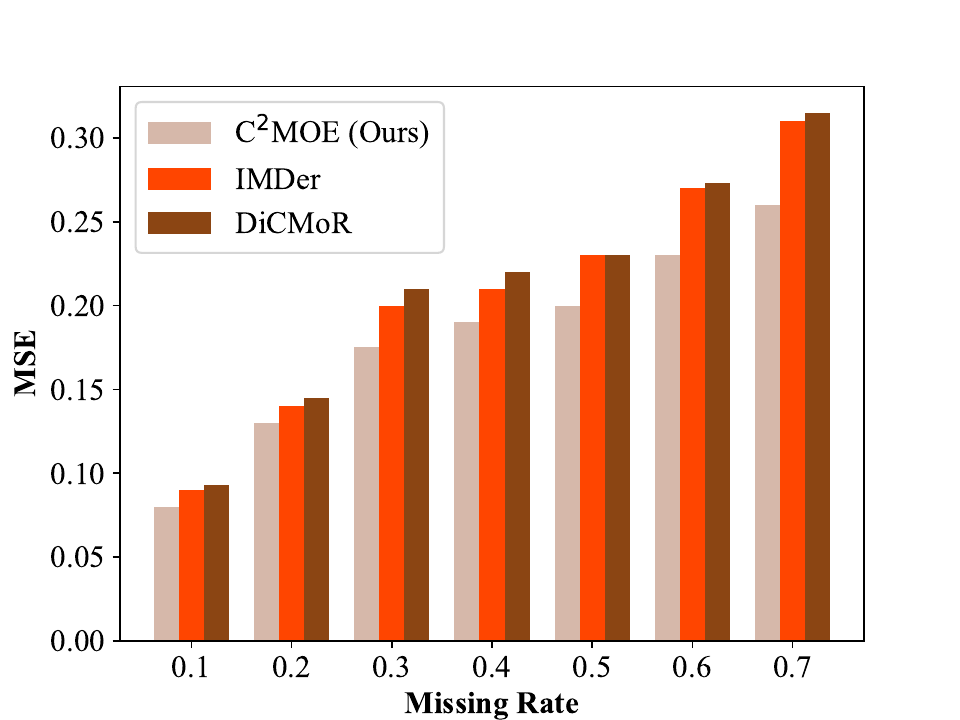}
		\label{fig:mosei}}
	\caption{The comparison of interpolation performance under different missing rates shows the interpolation effects of various methods when dealing with different missing rates.}
	\label{fig.1}
	\vspace{-3mm}
\end{figure}

\subsection{Visualization of Embedding Space}


Fig.~\ref{fig:enter-label} visualizes the feature distributions of the restored data and the original data under fixed missing-modality settings using t-SNE on the CMU-MOSEI dataset. As illustrated in the figure, the distributions generated by C\textsuperscript{2}MOE exhibit the closest alignment with those of the original data, indicating that our method is able to more effectively preserve the intrinsic structural and semantic characteristics of the missing modality during restoration. The restored features not only overlap well with the original distribution globally, but also maintain coherent local neighborhood structures. In contrast, other restoration-based methods show noticeable distributional shifts, with evident deviations and fragmented clusters—particularly in local regions where the overlap with the original data is weak. These discrepancies suggest that such methods struggle to maintain fine-grained semantic consistency, leading to distorted feature representations. Overall, this visualization provides intuitive evidence that C\textsuperscript{2}MOE better captures both global distributional consistency and local semantic structure, thereby achieving more faithful and robust modality restoration.

\subsection{Imputation Performance}

Fig. \ref{fig.1} reports the completion performance of different methods under varying missing rates. Across all missing-rate settings on CMU-MOSI and CMU-MOSEI, C$^2$MOE consistently outperforms the baselines, with especially notable gains at high missing rates. These results highlight the critical role of both distributional consistency and cross-modal complementarity in completion, which most baselines underexploit. By jointly maximizing and minimizing mutual information for consistency and complementarity, respectively, C$^2$MOE not only recovers missing modalities but also better preserves the original structural and semantic properties, particularly under severe missing conditions.

\section{Conclusions}

In this work, we tackle a core obstacle in Multimodal Emotion Recognition in Conversations, namely the severe performance degradation caused by missing modalities in realistic settings. We propose C²MOE, a Consistency and Complementarity guided Mixture of Experts framework that explicitly disentangles multimodal representations into consistency and complementarity components within a unified information theoretic formulation. A dual branch imputation mechanism couples uncertainty minimized consistency alignment with entropy maximized complementarity exploitation, enabling coherent reconstruction while preserving modality unique cues. In addition, a learnable reweighting module adaptively fuses expert outputs, improving flexibility under heterogeneous missing patterns and enhancing robustness. Extensive experiments on multiple MERC benchmarks show that C²MOE consistently surpasses state of the art methods across diverse missing modality scenarios, delivering superior accuracy, and stability. 
	
\bibliographystyle{ACM-Reference-Format}
\bibliography{refs}
	
\end{document}